\documentclass[10pt,twocolumn,letterpaper]{article}

\usepackage{wacv}
\usepackage{times}
\usepackage{epsfig}
\usepackage{graphicx}
\usepackage{amsmath}
\usepackage{amssymb}
\usepackage{booktabs}

\usepackage{subcaption}
\usepackage{graphicx}
\usepackage{float}
\usepackage{amsmath}
\usepackage{amssymb}
\usepackage{lipsum}
\usepackage{tabularx}
\usepackage{amsfonts}       
\usepackage{nicefrac}
\usepackage{microtype}
\usepackage{xcolor}
\usepackage{authblk}

\definecolor{SM_color}{rgb}{0.016, 0.600, 0.831}

\DeclareMathOperator*{\argmin}{arg\,min}

\def\wacvPaperID{700}
\wacvalgorithmstrack
\wacvfinalcopy

\ifwacvfinal
\usepackage[breaklinks=true,bookmarks=false]{hyperref}
\else
\usepackage[pagebackref=true,breaklinks=true,colorlinks,bookmarks=false]{hyperref}
\fi

\begin{document}

\title{Can Shadows Reveal Biometric Information?}

\author[]{Safa C. Medin}
\author[]{Amir Weiss}
\author[]{Fr\'edo Durand}
\author[]{William T. Freeman}
\author[]{Gregory W. Wornell}

\affil[]{Massachusetts Institute of Technology} 
\affil[]{\tt\small \vspace*{-0.2cm} \{medin, amirwei, fredo, billf, gww\}@mit.edu \vspace*{-0.1cm}}

\maketitle
\thispagestyle{empty}

\begin{abstract}
We study the problem of extracting biometric information of individuals by looking at shadows of objects cast on diffuse surfaces. We show that the biometric information leakage from shadows can be sufficient for reliable identity inference under representative scenarios via a maximum likelihood analysis. We then develop a learning-based method that demonstrates this phenomenon in real settings, exploiting the subtle cues in the shadows that are the source of the leakage without requiring any labeled real data. In particular, our approach relies on building synthetic scenes composed of 3D face models obtained from a single photograph of each identity. We transfer what we learn from the synthetic data to the real data using domain adaptation in a completely unsupervised way. Our model is able to generalize well to the real domain and is robust to several variations in the scenes. We report high classification accuracies in an identity classification task that takes place in a scene with unknown geometry and occluding objects.
\end{abstract}

\section{Introduction}
\label{sec:introduction}
\let\thefootnote\relax\footnotetext{\scriptsize This work was supported, in part, by NSF under Grant No.\ CCF-1816209 and the MIT-IBM Watson AI Lab under Agreement No.\ W1771646.}

\begin{figure}[t]
  \centering
   \includegraphics[width=0.90\linewidth]{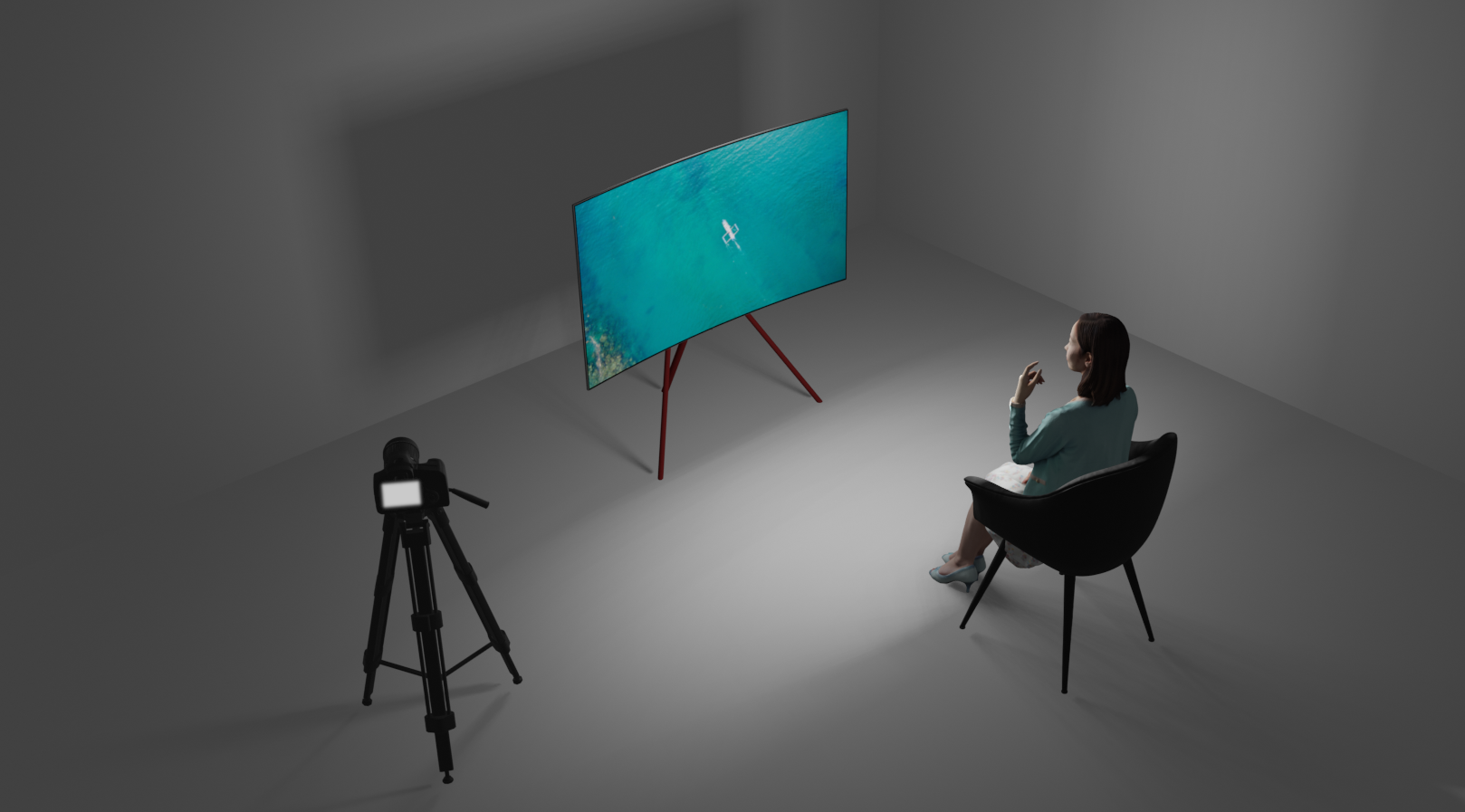}
   \caption{\small Consider an illustrative scenario where an individual sits across a TV screen in a given room, and suppose that the light reflected by the individual creates a shadow of the screen cast on a blank wall. We demonstrate that such shadows have a potential to leak biometric information in various scene configurations.
   }
   \label{fig:teaser}
\end{figure}

Imaging scenes that are not in our direct line-of-sight, referred to as non-line-of-sight (NLOS) imaging, has a diverse set of applications in several domains such as surveillance, search-and-rescue, robotic vision, and medical imaging~\cite{faccio2020non}. NLOS imaging methods typically aim to extract information from the hidden scenes that are outside of our field of view based on the observations of a visible scene. In this work, we ask the question whether or not such observations can leak \emph{sensitive} information about the hidden scenes. In particular, we introduce a novel problem where we seek to determine whether it is possible to extract \emph{biometric} information of individuals present in a room by looking at the shadows on diffuse surfaces induced by their presence as shown in Figure~\ref{fig:teaser}. We investigate this problem in a \emph{passive} NLOS imaging setting, meaning that we focus on scenarios where we rely on light sources naturally present in the scene.

Passive NLOS imaging methods can address a number of tasks such as recovering 2D images of the scene~\cite{saunders2019computational, yedidia2019using}, reconstructing videos of unknown scenes~\cite{aittala2019computational}, and estimating the motion and the number of hidden objects~\cite{bouman2017turning}. While several methods aim to recover the entire hidden scene~\cite{saunders2019computational, yedidia2019using, aittala2019computational}, often in accidental scenarios~\cite{torralba2012accidental} where no prior assumptions can be made about the scenes, recovering certain \emph{attributes} of the scene in such scenarios can be useful in several applications. For instance, 
determining whether or not a non-visible scene includes a person could be potentially useful for autonomous driving, security, or search-and-rescue  applications~\cite{rapp2020advances, sharma2021you}. Our focus, on the other hand, involves recovering biometric information of hidden individuals rather than merely detecting their presence. Here, we define the biometric information as any information that might be used to reveal an individual's identity, in whole or in part.

In this work, we aim to understand whether otherwise innocuous shadows can be used to reveal at least some biometric information by relying on existing contemporary learning tools. We approach this objective by focusing on a specific instance of biometric information extraction, namely, identity classification. In other words, we study the problem of recovering the identities of people in a given room by observing shadows of objects cast on a diffuse surface such as a blank wall. We emphasize that our approach does \emph{not} focus on shadows cast by the individuals but the shadows cast by the objects. We first carry out a maximum likelihood (ML) analysis of this task, characterizing performance under varying number of identities and noise levels by leveraging synthetic face data. Our results suggest that the information leakage from shadows can be significant under representative scenarios. Next, we investigate whether our findings are accurate predictions of what an adversary may be able to accomplish in practice by developing a learning-based method that discovers hidden biometric cues in the shadows without relying on any labeled real data. In particular, we build synthetic scenes composed of 3D face models obtained from a single photograph of each identity of interest~\cite{deng2019accurate}, and we transfer what we learn from this data to the real data in a completely unsupervised way by leveraging unsupervised domain adaptation techniques.  Our method generalizes well to the real domain and is robust to several variations in the scene, such as the shape of occluding objects, lighting conditions, head poses, and facial expressions. We report high classification accuracies in an identity classification task that takes place in a scene with unknown geometry and occluding objects, suggesting that, indeed, there is a significant biometric leakage phenomenon.

This work can be viewed as a first step towards understanding the degree to which seemingly benign images of shadow phenomema have the potential to leak at least some biometric information that could be of societal concern. Such information leakage might potentially be used with malicious intent, e.g., to determine the presence of an individual in a room without their consent. We emphasize that we deliberately do not seek to design an optimized identity classification system (such as a sophisticated adversary might want to). Rather, our methodology serves to demonstrate and characterize the biometric information leakage phenomenon to raise awareness to an overlooked privacy concern. Our results suggest that the biometric cues we discover in shadows could be used to distinguish identities as well as to reliably narrow the identity to within a group of individuals by extracting some amount of biometric information.

The main contributions of this work are as follows:
\begin{itemize}
    \item We introduce a timely biometric leakage question, which we formulate as a novel NLOS imaging problem of extracting an individual's identity from subtle, indirect shadow phenomena.
    \item Via a maximum likelihood analyis, we show that such shadows have a significant potential to leak sensitive information under representative scenarios.
    \item By combining existing learning tools, we develop a methodology that discovers biometric cues in the shadows without relying on any labeled real data, and report nontrivial accuracies in an identity classification task that takes place in a scene with unknown geometry and occluding objects.
\end{itemize}

\section{Background and Related Work}

We now  summarize the key background concepts and methodologies from 3D face modeling in computer graphics and domain adaption in machine learning, which we will leverage in our work.  We also include a brief summary of related work within NLOS imaging---albeit with different objectives---as  additional context for our contributions.

\textbf{Non-line-of-sight imaging.} Based on how the observed data is collected, NLOS imaging methods can be divided into two categories: \emph{active methods}, which typically involve an imaging device that consists of a coherent illumination source (laser) and a photon detector, and \emph{passive methods}, which do not require such specialized equipment. Passive methods have been explored in a variety of scene configurations and imaging objectives, and they typically exploit structure present in the scenes that induces \emph{occlusion}, which improves the conditioning of the imaging problem~\cite{yedidia2018analysis}. Among these methods, Bouman et al.~\cite{bouman2017turning} shows that vertical occluder structure such as corners can be used to recover 1D projection of a moving scene, from which the number of people moving in the hidden scene, their sizes and speeds can be estimated. Seidel et al.\ extend this idea to image stationary objects and form 2D reconstructions of the hidden scenes~\cite{seidel2019corner,seidel2020two}, while Naser et al.~\cite{naser2018shadowcam} detects obstacles around the corners for autonomous driving applications. 
However, none of these methods explore extracting biometric information from NLOS measurements. In a different setting,~\cite{saunders2019computational} uses a pinspeck occluder with known shape but unknown position to recover 2D scenes while~\cite{yedidia2019using} exploits motion in hidden scenes to recover the scene without any assumptions about the occluder shape and position. More recently,~\cite{wang2021accurate} and ~\cite{sharma2021you} study classification tasks from NLOS measurements. Unlike our more practical \emph{unsupervised} approach, however, these methods use supervised learning tools and do not focus on identity classification. In active imaging methods, on the other hand, several patches of the observed scene are illuminated so that the light pulses reflecting on these patches reach the hidden scene and are reflected back to the photon detector through the observed scene. The increasing availability of less expensive time-of-flight sensors has enabled the proliferation of active NLOS imaging methods~\cite{ buttafava2015non, o2018confocal, rapp2017few, heide2019non, rapp2020seeing, wu2021non}.

\textbf{3D morphable face models.} 
 3D morphable face models are statistical models of human faces~\cite{blanz1999morphable, paysan20093d,  FLAME:SiggraphAsia2017, gerig2018morphable}, which have been widely used in domains such as face recognition, entertainment, neuroscience and psychology~\cite{egger20203d}. Over the last decade, advances in deep learning allowed these models to achieve remarkable results in the challenging problem of recovering 3D faces from 2D images~\cite{richardson20163d, dou2017end, tuan2017regressing, richardson2017learning}, where some of the more recent approaches do not require explicit 3D shape labels~\cite{tewari2017mofa, tewari2018self, tran2018nonlinear, genova2018unsupervised, deng2019accurate}. Among these methods, Deng et al.~\cite{deng2019accurate} develops a reconstruction network that recovers accurate 3D faces from a single image, which we use in our synthetic data collection.

\textbf{Domain adaptation.} Over the last few years, there has been a significant amount of work in the area of domain adaptation~\cite{wang2018deep, wilson2020survey}, which is the study of transferring knowledge learned from a source domain to a target domain. More recent approaches in domain adaptation are more concentrated towards deep learning-based solutions and unsupervised methods where no labels from the target domain are used. These methods commonly rely on aligning the distributions of the source and target domains in feature spaces~\cite{ganin2015unsupervised, tzeng2015simultaneous, tzeng2014deep, long2015learning, sun2016deep, liu2016coupled, tzeng2017adversarial, french2017self}. Among these methods, DDC~\cite{tzeng2014deep} aims for learning domain-invariant representations by imposing a maximum mean discrepancy loss~\cite{gretton2009covariate}, Deep CORAL~\cite{sun2016deep} aligns the second-order statistics of the source and the target domains, while ADDA~\cite{tzeng2017adversarial} employs an adversarial discriminator to make the representations of the two domains indistinguishable from each other. In another approach, Li et al.~\cite{li2018adaptive} shows that updating the batch normalization statistics~\cite{ioffe2015batch} for the target domain can also be very effective, which we employ in our method.

\section{Methodology}

Suppose we are given $M$ different identities who are individually present in a room with an unknown geometry, and suppose we
observe shadows cast by an occluder in the room blocking the light reflected
by each individual. Denoting each observation as $\mathbf{x} \in \mathbb{R}^{n}$ (grayscale images of resolution $\sqrt{n} \times \sqrt{n}$, with $\sqrt{n}\in\mathbb{N} $) and its ground truth identity label as $y \in \mathcal{Y} = \{1, 2, \dots, M\}$, our objective is to learn a classifier that is capable of reliably inferring identities from shadows given training data $\mathcal{S} = \{(\mathbf{x}_1, y_1), \dots, (\mathbf{x}_N, y_N) \}$. 

We begin presenting our approach by first describing our 3D face model. Then, we follow an ML analysis to understand how much information is leaked by shadows under varying numbers of identities and noise levels. Inspired by our findings, we develop our learning-based methodology that could be employed to distinguish identities in practice.

\label{sec:formulation}

\subsection{3D Face Modeling}

In our approach, we make use of synthetic face models which represent faces as triangular meshes. Given a number of vertices $V\!$ in a mesh, we represent a face shape $\mathbf{s} \in \mathbb{R}^{3V}$ (3D coordinates for each vertex) and its texture $\mathbf{t} \in \mathbb{R}^{3V}$ (RGB colors for each vertex) with the following linear 3D morphable model~\cite{paysan20093d, cao2013facewarehouse}:
\begin{equation}
    \begin{aligned}
     \mathbf{s} &= \mathbf{\bar{s}} + \mathbf{M}_{\mathrm{id}} \boldsymbol{\alpha}_{\mathrm{id}} + \mathbf{M}_{\mathrm{exp}} \boldsymbol{\alpha}_{\mathrm{exp}} \\
    \mathbf{t} &= \mathbf{\bar{t}} + \mathbf{M}_{\mathrm{tex}} \boldsymbol{\alpha}_{\mathrm{tex}}
    \label{eq:3dmm}
\end{aligned}
\end{equation}
where \smash{$\mathbf{\bar{s}} \in \mathbb{R}^{3V}$} and \smash{$\mathbf{\bar{t}} \in \mathbb{R}^{3V}$} are the mean shape and mean texture of the model; \smash{$\mathbf{M}_{\mathrm{id}} \in \mathbb{R}^{3V\times k_\mathrm{id}} $}, $\mathbf{M}_{\mathrm{exp}} \in \mathbb{R}^{3V\times k_\mathrm{exp}}$ and $\mathbf{M}_{\mathrm{tex}} \in \mathbb{R}^{3V\times k_\mathrm{tex}}$ are the identity, expression and texture bases; and \smash{$\boldsymbol{\alpha}_\mathrm{id} \in \mathbb{R}^{k_\mathrm{id}}$}, \smash{$\boldsymbol{\alpha}_\mathrm{exp} \in \mathbb{R}^{k_\mathrm{exp}}$} and \smash{$\boldsymbol{\alpha}_{\mathrm{tex}}\in \mathbb{R}^{k_\mathrm{tex}}$} are the identity, expression and texture coefficients. Here, \smash{$\mathbf{\bar{s}}, \mathbf{\bar{t}}, \mathbf{M}_{\mathrm{id}}, \mathbf{M}_{\mathrm{exp}}$}, and \smash{ $\mathbf{M}_{\mathrm{tex}}$} are all provided by the model, whereas \smash{$\boldsymbol{\alpha}_\mathrm{id}$} and \smash{$\boldsymbol{\alpha}_{\mathrm{tex}}$} are fixed and known vectors obtained via sampling from a Gaussian prior~\cite{egger20203d} or by reconstructing 3D faces from 2D images of the identities of interest.

\subsection{Maximum Likelihood Analysis}

\label{sec:ml-approach}

To determine whether and how much biometric information leaks from the shadows, we simulate a representative scene in the synthetic domain. For this, we first describe our data generation, and in particular the creation of 3D faces using the model described in~\eqref{eq:3dmm} and the convolutional model of occlusion. Next, we present our ML-based learning algorithm with an objective to obtain lower bound on the classification accuracy with respect to numbers of identities and noise levels, where we make certain assumptions regarding the data distribution to allow for ease of analysis.

\subsubsection{Observation Model}

Let \smash{ $(\boldsymbol{\alpha}_\mathrm{id}^{m}, \boldsymbol{\alpha}_\mathrm{tex}^{m}), \, m=1,2,\dots, M$}, denote \emph{fixed} and \emph{known} identity and texture coefficients of $M$ identities. For simplicity, suppose that the faces are sufficiently far away from the observation surface so that they can be represented as 2D rendered images of the 3D face objects, and suppose that we observe grayscale images of shadows, $\mathbf{x}^m \in \mathbb{R}^n$ for each identity $m$, according to the following data model:
\begin{equation}
\mathbf{x}^{m} =
\mathbf{A} \underbrace{\mathbf{R}^{m} \widetilde{\mathbf{M}}_{\mathrm{tex}} \boldsymbol{\alpha}_\mathrm{tex}^{m}}_{\triangleq \ \mathbf{r}^m} + \mathbf{z}=\mathbf{A}\mathbf{r}^{m}+\mathbf{z}, \quad m=1,\dots, M
\label{eq:data-model}
\end{equation}
where $\mathbf{\widetilde{M}}_{\mathrm{tex}} \triangleq [\mathbf{M}_{\mathrm{tex}} \ \ \mathbf{\bar{t}}\,] \in \mathbb{R}^{3V \times (k_\mathrm{{tex}}+1)}$ is the augmented texture basis that generates a texture map from a texture code \smash{$\boldsymbol{\alpha}_\mathrm{tex}^{m} \in \mathbb{R}^{(k_\mathrm{{tex}}+1)}$}; the random matrix \smash{$\mathbf{R}^{m}\triangleq \mathbf{R}(\boldsymbol{\alpha}_\mathrm{id}^{m}, \boldsymbol{\alpha}_\mathrm{exp}^{m}, \boldsymbol{\theta}^{m}, \boldsymbol{\gamma}^{m} ) \in \mathbb{R}^{n\times 3V}$} denotes the rendering operation that maps vertex colors of the mesh to grayscale image pixels as a deterministic {nonlinear} function of the {fixed} and {known} identity vector \smash{$\boldsymbol{\alpha}_\mathrm{id}^{m} \in \mathbb{R}^{k_\mathrm{{id}}}$}, {random} expression vector \smash{$\boldsymbol{\alpha}_\mathrm{exp}^{m} \in \mathbb{R}^{k_\mathrm{{exp}}}$}, {random} pose vector \smash{$\boldsymbol{\theta}^{m} \in \mathbb{R}^{k_\mathrm{{\theta}}}$} and {random} lighting \smash{$\boldsymbol{\gamma}^{m} \in \mathbb{R}^{k_\mathrm{{\gamma}}}$}; \smash{$\mathbf{A} \in \mathbb{R}^{n\times n}$} is an {unknown} light transport matrix that maps a face image to a shadow image due to the presence of an occluder; and finally $\mathbf{z} \sim \mathcal{N}(\mathbf{0}_n,\sigma^2 \mathbf{I}_n) \in \mathbb{R}^n$ denotes the additive noise due to thermal noise and shot noise~\cite{yedidia2018analysis}, assumed to be statistically independent of the signal-related term $\mathbf{A}\mathbf{r}^m$, where $\mathbf{I}_n$ is the $n$-dimensional identity matrix.

\subsubsection{3D Face Generation}

\begin{figure}[t]
    \centering
    \includegraphics[width=\linewidth]{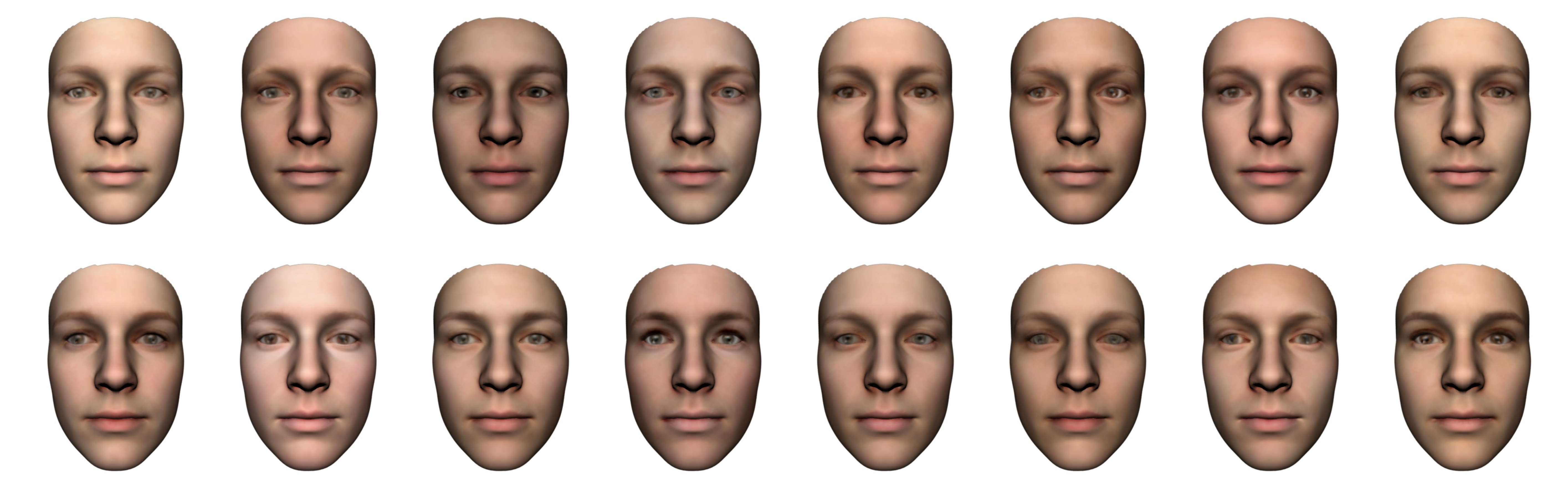}
    \caption{\small $M = 16$ randomly sampled identities we use in our experiments, rendered with neutral expressions (i.e., $\boldsymbol{\alpha}_\mathrm{exp} = \mathbf{0}$), same head poses and lighting conditions. Identities differ only with respect to their textures. In the experiments, textures are converted to grayscale to avoid potential reliance on color information.}
    \label{fig:random-ids}
    \vspace{-1mm}
\end{figure}

We generate faces using the Basel Face Model 2009~\cite{paysan20093d} with the neck and the ear regions excluded. We randomly generate $M = 16$ identities by sampling $\boldsymbol{\alpha}_{\mathrm{tex}}$ from $\mathcal{N}(\mathbf{0}_{k_{\mathrm{tex}}}, \mathbf{I}_{k_{\mathrm{tex}}})$ and
setting $\boldsymbol{\alpha}_{\mathrm{id}} = \mathbf{0}_{k_{\mathrm{id}}}$, i.e., the identities have the same face shape under the same facial expression. Hence, the differences between identities only result from the texture, as we seek to show that subtle differences in texture alone can be sufficient to distinguish identities from each other. The expression basis is provided by the model constructed from the FaceWarehouse dataset \cite{cao2013facewarehouse}, which we use to sample identities with varying expressions. Finally, we render the faces with random pose and lighting parameters ($\boldsymbol{\theta}, \boldsymbol{\gamma}$). We illustrate the 16 identities under neutral expressions, same head poses, and lighting conditions in Figure~\ref{fig:random-ids}. \vspace{-2mm}

\begin{figure}[t]
    \centering
    \includegraphics[width=\linewidth]{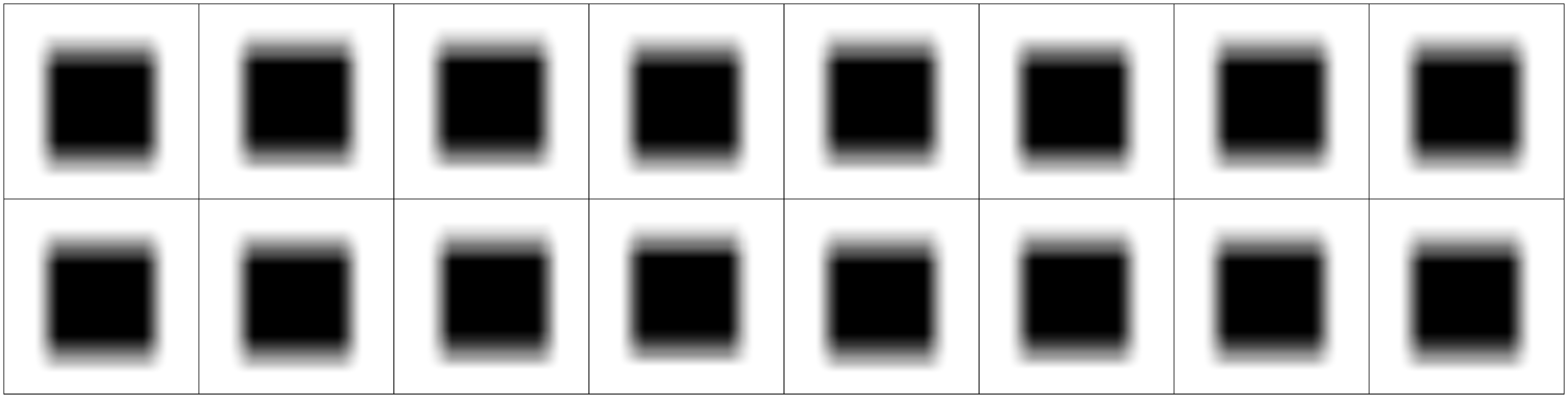}
    \caption{\small Representative shadow images for each identity, normalized to the range $[0,1]$ for illustration purposes.}
    \label{fig:shadows-conv}
    \vspace{-2mm}
\end{figure}

\subsubsection{Convolutional Model of Occlusion}

In this part of our analysis only, we assume that the face and the occluder lie in 2D planes that are parallel to each other as well as to the observation plane. This gives rise to the convolutional model of occlusion, a model commonly adopted in passive NLOS imaging applications~\cite{krishnan2011blind, yedidia2019using}. Under this model, assuming the occluder is an opaque object that completely blocks the light, the light transport matrix $\mathbf{A}$ can be defined as $\mathbf{A}_{ij} = 0$ if the occluder blocks the light coming from $i$-th pixel of the face image to $j$-th pixel of the observation plane, and $\mathbf{A}_{ij} = 1/n$ otherwise. Here, the $(1/n)$-scaling ensures that the observed total power does not exceed the total power reflected from the face~\cite{yedidia2018analysis, ajjanagadde2019near}. We illustrate representative images of shadows obtained with this model in Figure~\ref{fig:shadows-conv}, where a rectangular occluder is simulated.

\subsubsection{Learning Algorithm}
\label{sec:ml-algorithm}

Let $\mathbf{r}^{m}$ be the random vector representing random face images for identity $m$ as defined in~\eqref{eq:data-model}. Given a fixed but unknown $\mathbf{A}$, we assume that the \emph{noiseless} images of shadows $\mathbf{A} \mathbf{r}^{m}$ are normally distributed with mean $\boldsymbol{\mu}^{m}$ and covariance $\mathbf{\Sigma}^{m}$, i.e., $\mathbf{A} \mathbf{r}^{m} \sim \mathcal{N}(\boldsymbol{\mu}^{m}, \mathbf{\Sigma}^{m})$. Since we assume that the noise $\mathbf{z}$ is statistically independent of these images, the labeled examples $\mathbf{x}^{m}$ for each identity $m$ are distributed according to $\mathcal{N}(\boldsymbol{\mu}^{m}, \mathbf{Q}^m)$, where $\mathbf{Q}^m \triangleq \mathbf{\Sigma}^{m} + \sigma^2 \mathbf{I}_n$. Given training data $\{\mathbf{x}_1^{m}, \dots, \mathbf{x}_N^{m} \}$ for each identity $m$, we first compute the sample means and covariances as
\begin{equation}
    \boldsymbol{\widehat{\mu}}^{m} \triangleq \frac{1}{N}\sum_{i=1}^N \mathbf{x}_i^{m} \quad \mathbf{\widehat{Q}}^{m} \triangleq \frac{1}{N} \sum_{i=1}^N  (\mathbf{x}_i^{m} -  \boldsymbol{\widehat{\mu}}^{m}) (\mathbf{x}_i^{m} -  \boldsymbol{\widehat{\mu}}^{m})^\mathrm{T}
\end{equation}
At test time, given a test observation $\mathbf{x}$, assuming each identity is equally likely and that the determinant of the covariance matrices of each identity are equal to each other, the ML estimation rule is given by\footnote{In fact, this is only \emph{asymptotically} an ML classifier, since we use the sample means and covariances obtained from a finite number of samples.}
\begin{equation}
    \widehat{m} = \argmin_{m=1,2,\dots,M} (\mathbf{x} - \boldsymbol{\widehat{\mu}}^{m})^\mathrm{T} \big(\mathbf{\widehat{Q}}^{m}\big)^{-1}
    (\mathbf{x} - \boldsymbol{\widehat{\mu}}^{m})
    \label{eq:ml-estimator}
\end{equation}
In practice, since $n$ is typically very large, we cannot assume that $N \gg n$. Therefore, inverting the sample covariance matrices obtained by a finite number of samples $N$ does not yield a robust classifier. Assuming $\mathrm{rank}(\mathbf{\Sigma}^m) = r$ for any $m$, and denoting the eigenvalue decomposition of the $m$-th sample covariance matrix $\mathbf{\widehat{Q}}^{m} \triangleq \mathbf{U}^m \mathbf{\Lambda}^m (\mathbf{U}^m)^\mathrm{T}$ with $\mathbf{\Lambda}^m = \mathrm{diag}(\lambda_1^m, \dots, \lambda_n^m)$ such that $\lambda_1^m \geqslant \dots \geqslant \lambda_n^m > 0$, we propose the following improved procedure. Since the noise is assumed to be spatially white, we first estimate the noise variance for each identity as~\cite{wax1985detection} \vspace{-0.2cm}
\begin{equation}
\widehat{\sigma}^2 = \frac{1}{n-r} \sum_{i=r+1}^{n} \lambda_i^m
\end{equation}
Then, we set the refined estimate $\mathbf{\widehat{Q}}^{m} =  \mathbf{U}^m \mathbf{\bar{\Lambda}}^m (\mathbf{U}^m)^\mathrm{T}$, where $\mathbf{\bar{\Lambda}}^m = \mathrm{diag}(\lambda_1^m, \dots, $ $\lambda_r^m, \widehat{\sigma}^2, \dots, \widehat{\sigma}^2)$, and adopt the ML estimator defined in \eqref{eq:ml-estimator}. 

\subsection{Neural Network Classifier}

As an example of the kind of processing an adversary seeking to extract biometric information might use in practice, we now develop a learning-based framework suitable for identity classification in real settings, where we assume that the occluder shape is not fixed but arbitrary. Since we follow a data-driven approach, representing possible variations such as the occluder shape, lighting conditions, facial expressions, and head poses in the training data is crucial to achieve a robust classification system. Because collecting such data is highly impractical, we develop a method that avoids such challenges. In particular, we use 3D graphics software to collect large amounts of training data by placing 3D faces and objects into simulated scenes. Then, we transfer what we learn from these scenes to the real settings by employing unsupervised domain adaptation techniques.

\begin{figure}[t]
    \centering
    \includegraphics[width=\linewidth]{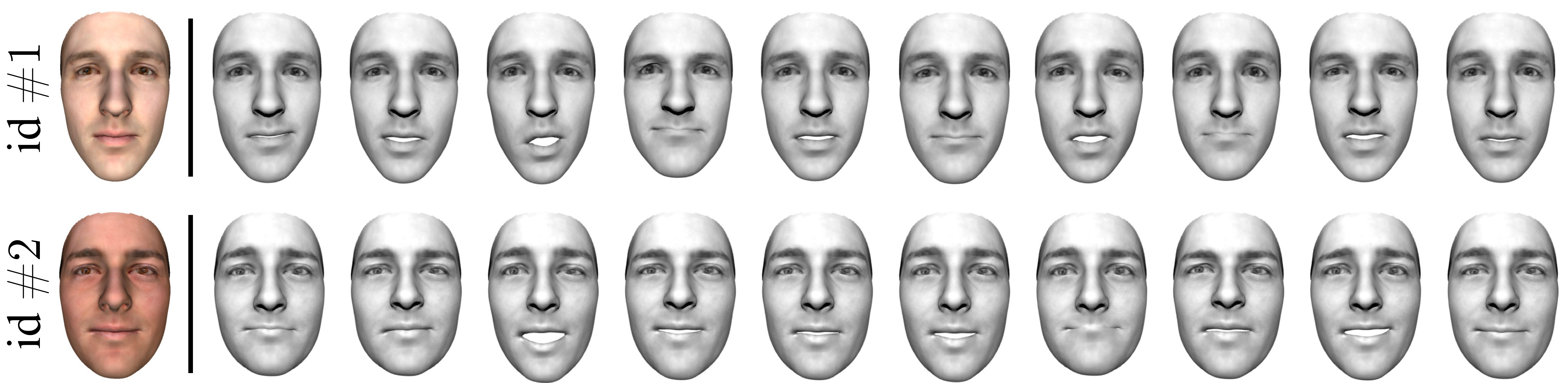}
    \caption{\small Face reconstructions of $M=2$ identities we use in our experiments, rendered with varying expressions. Given RGB reconstructions of the faces, we first convert their textures to grayscale and match their average intensity levels. Expressions are randomly sampled and varied in the dataset.}
    \label{fig:expression-samples}
    \vspace{-4mm}
\end{figure}

\subsubsection{3D Face Generation}

To minimize the gap between the synthetic and real domains, we use a 3D face reconstruction network~\cite{deng2019accurate}, which allows us to obtain a 3D model of an identity \emph{from a single image}. The reconstructed faces in this work also follow the Basel Face Model 2009 \cite{paysan20093d} with the neck and the ear regions excluded, which ensures that the network trained with the synthetic data relies only on the identity information, i.e.,\ information such as the thickness of the neck or the contrast between the hair and skin intensities cannot be exploited in our method. As before, we create variations in facial expressions using the model obtained from~\cite{cao2013facewarehouse}. We convert the reconstructed textures to grayscale to avoid potential reliance on color information, and scale the intensity levels of the two identities so that the average intensity of their textures are the same. We show the reconstructed faces with varying expressions in Figure~\ref{fig:expression-samples}. \vspace{-3mm}

\subsubsection{Scene Geometry and Datasets}

Our imaging configuration includes the following: a person whose identity is unknown, a light source that illuminates the face of this person, a blank wall where we make our observations, and an occluding object that creates shadows on this wall. For illustration purposes, we focus on \emph{chairs} as occluding objects, as they are one of the most common and diverse classes of indoor objects. However, we emphasize that our method can easily be extended to handle more classes of objects by incorporating them in the training set.

\begin{figure}[t]
\centering
    \begin{subfigure}[H]{0.85\columnwidth}
            \includegraphics[width=1.0\columnwidth]{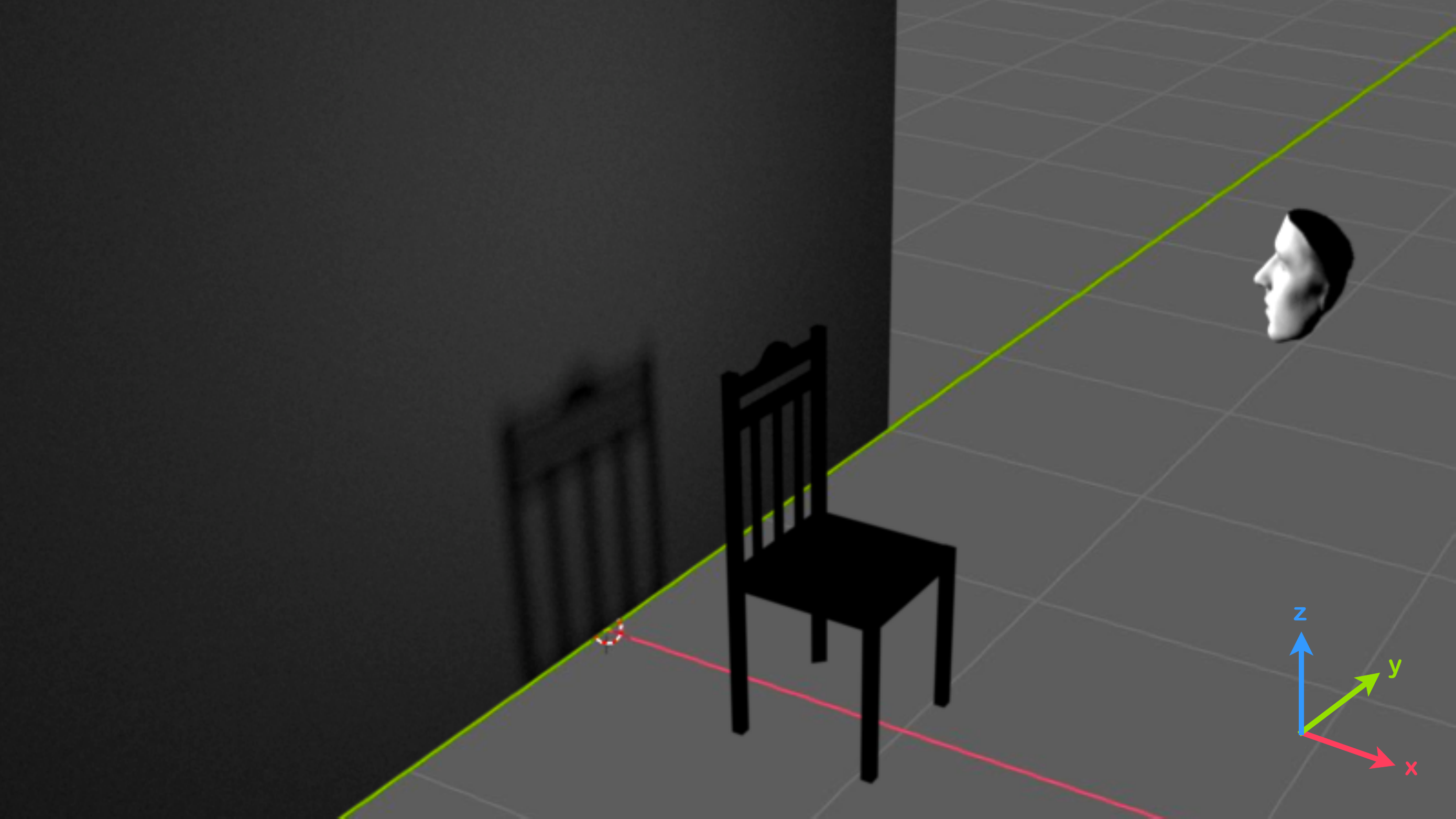}
            \caption{Synthetic scene geometry}
             \vspace{0.2cm}
            \label{fig:synth-geo}
    \end{subfigure}
    \begin{subfigure}[H]{0.85\columnwidth}
            \includegraphics[width=1.0\columnwidth]{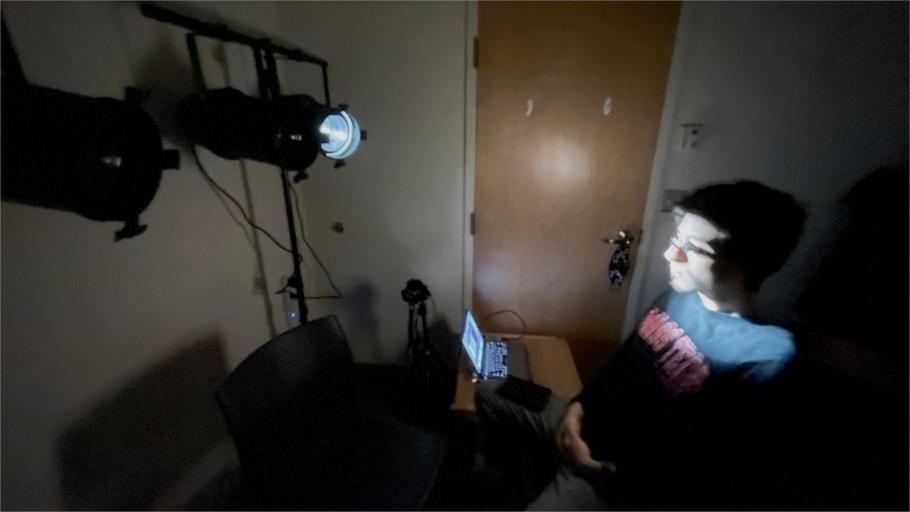}
            \caption{Real scene geometry}
            \label{fig:real-geo}
    \end{subfigure}
    \caption{\small Scene geometries for (a) synthetic and (b) real settings. Both scenes consist of four main components: a person whose identity is unknown, an illumination source, a blank wall, and an occluding object that creates the shadows on the wall. The light reflecting from the face creates shadows of objects on the wall.}
        \vspace{-3mm}
\end{figure}

In our synthetic data collection, we use 3D chair models provided by ShapeNet \cite{chang2015shapenet}. We use a white planar object as a wall and a white spotlight as an illumination source. When we render the scenes, we cover as much variation as possible by changing the pose, position and expression of the faces, and vary the position of the light sources. We illustrate a representative synthetic scene in Figure~\ref{fig:synth-geo}.

In our real data collection, the individuals sit across a blank wall individually, where a chair is positioned between the identity and the wall. The faces are illuminated by spotlights in different positions while the expressions and poses of the subjects, as well as the pose of the chair, are varied during the data collection process. We performed these experiments in a physical space shown in Figure~\ref{fig:real-geo}.

\subsubsection{Domain Adaptation}

Given two sets of data $\mathcal{S} = \{(\mathbf{x}_1^s, y_1^s), \dots, (\mathbf{x}_N^s, y_N^s) \}$ and $\mathcal{T} = \{(\mathbf{x}_1^t, y_1^t), \dots, (\mathbf{x}_N^t, y_{N}^t) \}$, which represent the source data and the target data, respectively, our objective is to learn a classifier using the source data $\mathcal{S}$ such that it performs well on the target data $\mathcal{T}$. This can be achieved in a supervised manner by using very few labeled samples from $\mathcal{T}$, or in an unsupervised manner by using no labeled samples from $\mathcal{T}$. In this work we follow the latter, as we seek to ensure that the supervision signals coming from the target domain involve only identity information, i.e.,\ these signals may depend on unintended cues from the real settings such as clothing, reflectance of the hair or other unintended phenomena.

Our method involves training a classification network that follows the ResNet-18 architecture~\cite{he2016deep}, where we change the final classification layer so that it reflects the number of classes in our application. Initializing the feature extraction module with the pretrained weights, we first train the network on the synthetic data in a supervised manner. Then, we freeze the learned weights and update the running means and variances of each batch normalization layer in the network~\cite{ioffe2015batch, li2018adaptive} by feeding the unlabeled target data $\mathcal{T} = \{\mathbf{x}_1^t, \dots, \mathbf{x}_N^t \}$ through the network. As we will show in the next section, the updated network generalizes well to the test samples from the target domain.

\section{Experiments and Results}
\label{sec:experiments}

We describe our experiments and present their results by first focusing on our ML classifier and characterizing its performance with respect to the number of identities and noise levels. Next, we focus on our neural network classifier by elaborating on the real and synthetic data colletion, and provide accuracies obtained in different stages of the method.

\begin{figure}[t]
    \centering
    \includegraphics[width=0.80\columnwidth]{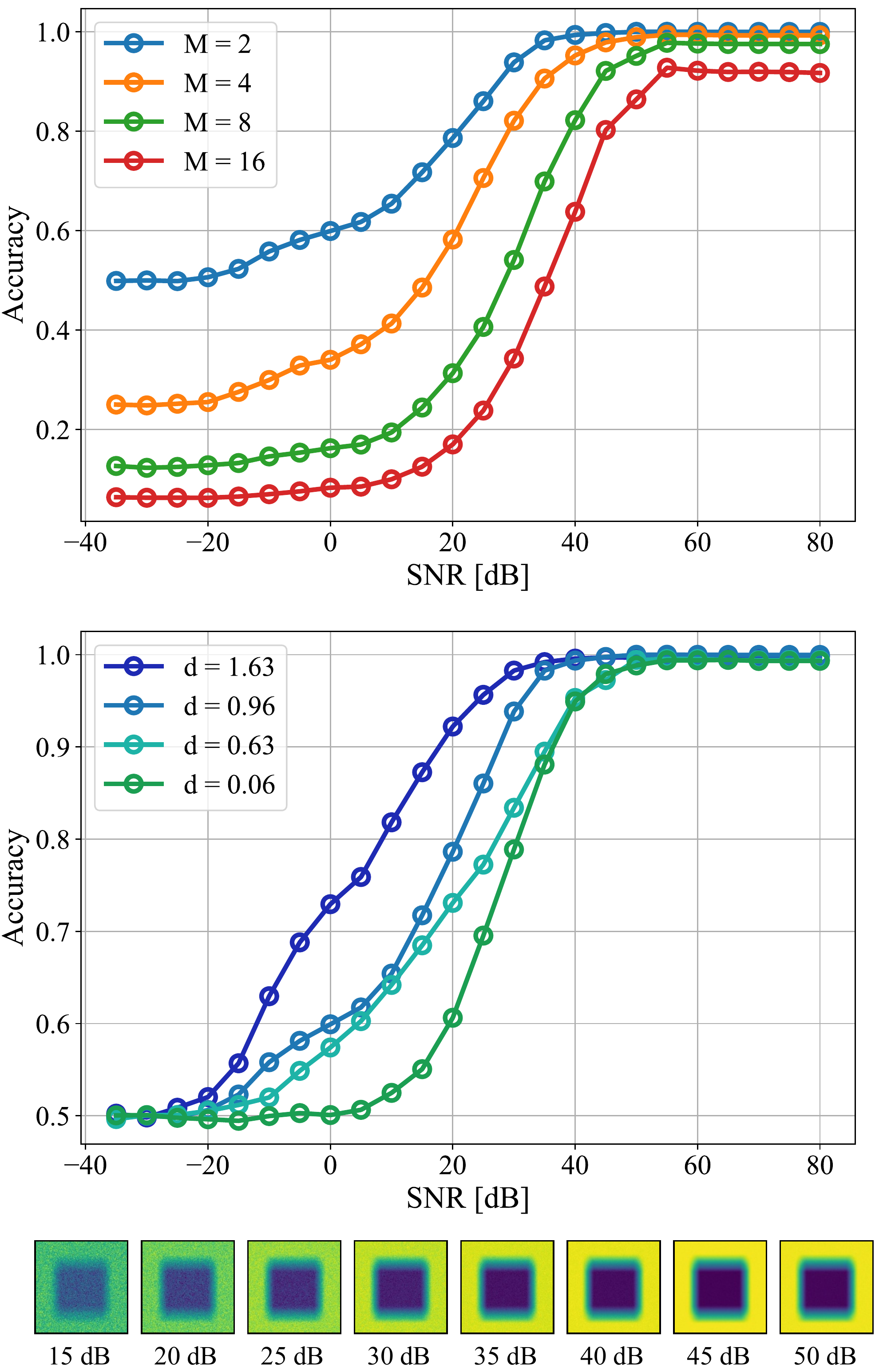}
    \caption{\small Empirical results for maximum likelihood classification. \emph{Top}: Classification accuracies with respect to varying number of identities and signal-to-noise ratio (SNR). \emph{Bottom}: Classification accuracy with respect to the distance between the estimated means of the identities and SNR.}
    \label{fig:accs-ml}
    \vspace{-4.5mm}
\end{figure}

\subsection{Maximum Likelihood Analysis}

Following the data generation pipeline we described in Section~\ref{sec:ml-approach}, we generate $20000$ samples for each of the $16$ identities, where we randomly change the head poses, facial expressions, and lighting conditions (we include the details of this process in the supplementary material). We split this data into train and tests sets with $90\%$--$10\%$ split, which gives us $18000$ train and $2000$ test samples for each identity. Next, we pick the first 2, 4, 8, and then all 16 identities from this dataset, and apply our ML-based learning algorithm described in Section~\ref{sec:ml-algorithm}, where we calculate classification accuracies on the test data under varying noise levels. We summarize our results in Figure~\ref{fig:accs-ml} (\emph{top}), where we observe high accuracies for all numbers of identities at moderate-to-high signal-to-noise ratio (SNR) levels.

Although the results we illustrate in Figure~\ref{fig:accs-ml} (\emph{top}) are representative under the face model described in Section~\ref{sec:formulation}, we note that the similarities between the identities have a natural influence on the accuracy, since it is more difficult to distinguish two identities that have very similar face shapes and textures (e.g., two identities might include identical twins in practice). To investigate this, we pick $4$ pairs of identities from our dataset with varying distances $d(i,j)$ between their estimated means, where $d(i,j) \triangleq \Vert \boldsymbol{\hat{\mu}}^i - \boldsymbol{\hat{\mu}}^j \Vert_2  \in \mathbb{R}_+$ for an identity pair $(i,j)$. We report accuracy curves for these pairs in Figure~\ref{fig:accs-ml} (\emph{bottom}), where we observe that the distance between the means has a direct impact on the performance, although almost perfect classification is still possible when the SNR is sufficiently high. This suggests that the variabilities of shadow images associated with each identity (determined by the respective covariances) are almost negligible with respect to the distances between the means when the noise is sufficiently small.

\begin{figure}[t]
    \centering
    \includegraphics[width=\linewidth]{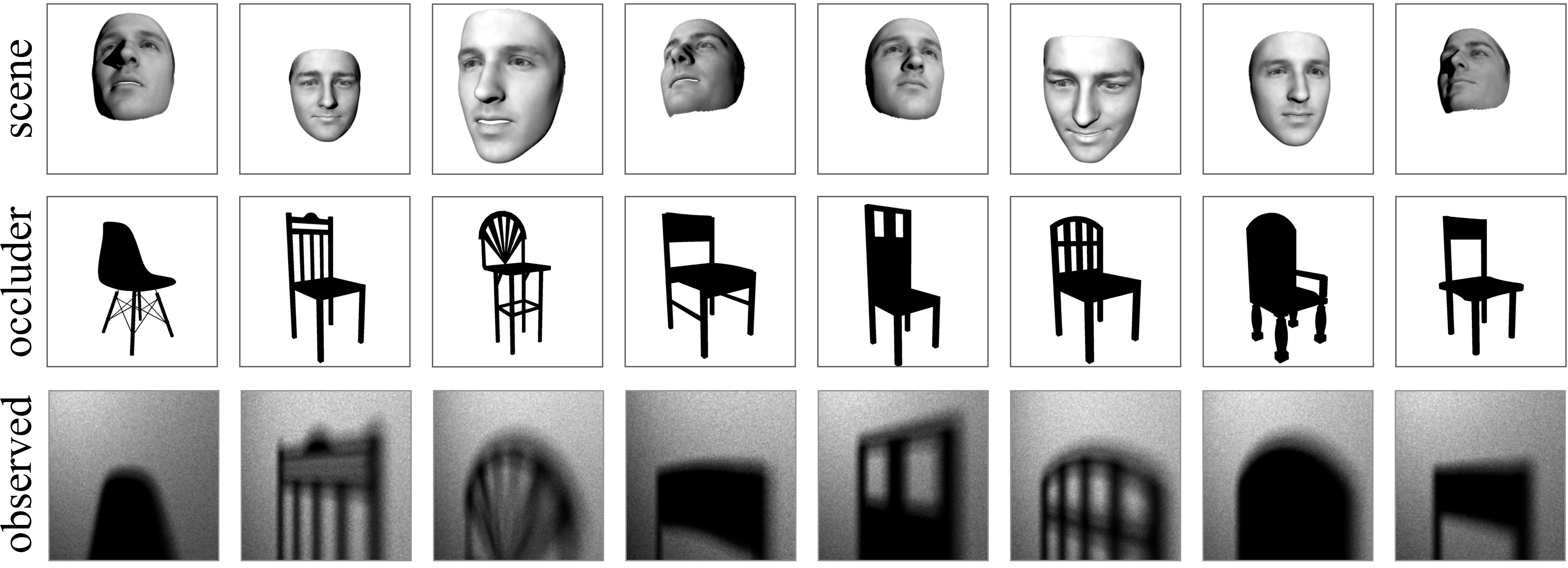}
    \caption{\small Representative samples from the dataset, where each column shows one sample. Our dataset covers a diverse set of head poses and facial expressions as well as occluder shapes.}
    \label{fig:dataset-10-samples}
\end{figure}

\subsection{Neural Network Classifier}
We now present our experiments and results for the neural network classifier, and show that our method is effective in extracting subtle biometric cues from shadows in real settings, consistent with our ML analysis predictions.

\subsubsection{Synthetic Data Collection and Training}

We generate the synthetic data for our network randomly, where we vary the pose, expression and position of the face, the location of the light source, and the occluder shape. We illustrate representative samples from the dataset in Figure~\ref{fig:dataset-10-samples}.

\begin{figure}[t]
\centering
    \begin{subfigure}[H]{0.85\columnwidth}
            \includegraphics[width=1.0\columnwidth]{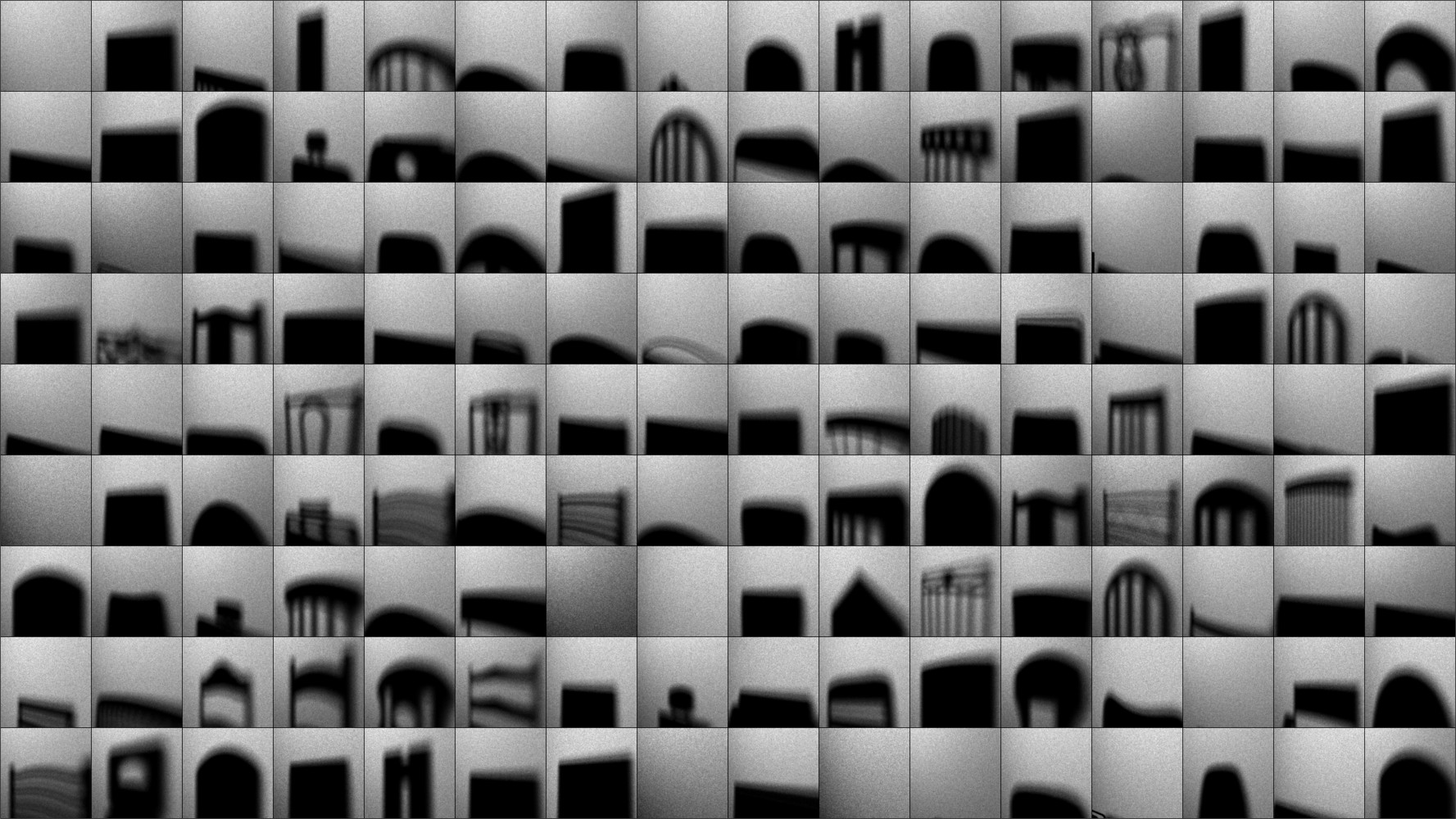}
            \caption{Source images}
            \vspace{0.2cm}
            \label{fig:source-images}
    \end{subfigure}
    
    \begin{subfigure}[H]{0.85\columnwidth}
            \includegraphics[width=1.0\columnwidth]{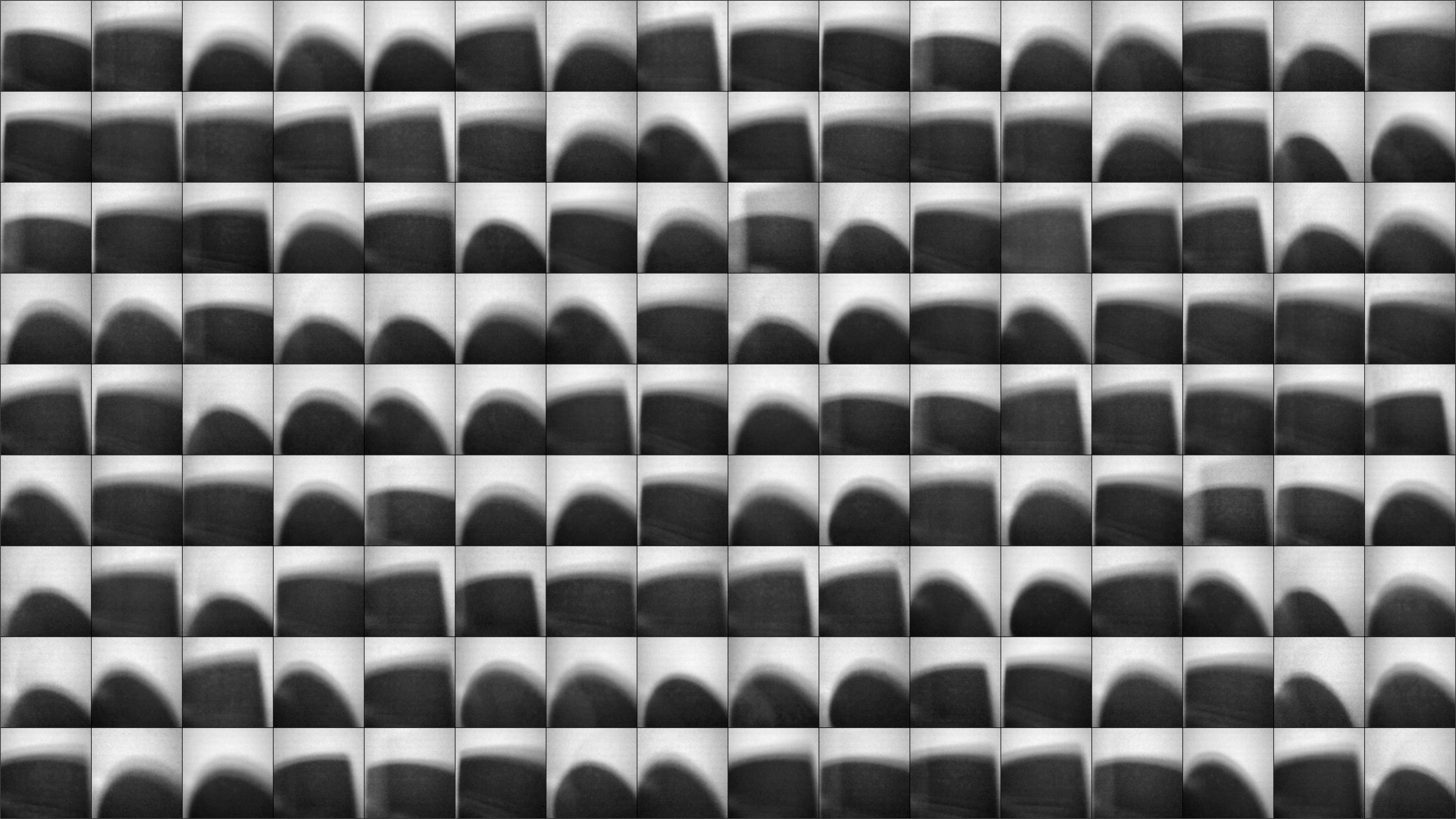}
            \caption{Target images}
            \label{fig:target-images}
    \end{subfigure}
    \vspace{-1mm}
    \caption{\small Random images from the source and the target datasets.}
    \vspace{-3mm}
\end{figure}

We collect our synthetic data using Mitsuba2 \cite{nimier2019mitsuba}, with which we render $256\times256$ images of the observed wall using $50000$ samples per pixel. Rendering one image takes $\approx50$ seconds on an NVIDIA GeForce RTX 2080 Ti GPU, and the intensities of all images are normalized to $[0,1]$ range after rendering. For each identity, we collect $4000$ images that we split into train and test sets with $75\%$--$25\%$ split, which gives us a total of $6000$ train and $2000$ test samples. We illustrate random samples from this dataset in Figure~\ref{fig:source-images}. Details of the dataset generation and training procedure are included in the supplementary material.

\subsubsection{Real Data Collection and Domain Adaptation}

To represent the typical use cases, we deliberately cover fewer variations in our real data compared to the synthetic data (e.g., collecting data in a very diverse set of scene configurations may not be feasible or practical for an adversary). In particular, we experiment with $4$ light source locations by using $4$ separate spotlights (which are individually lit during the data collection), and $2$ different occluders which we repose in $5$ different angles to increase the diversity in the dataset. Similar to what we have in the synthetic dataset, the identities also change their head poses and facial expressions while the data is collected. We collect $4000$ samples for each identity, and we randomly split the whole dataset into train and test sets with $75\%-25\%$ split. We illustrate random samples from the real dataset in Figure~\ref{fig:target-images}.

\begin{figure}[t]
\centering
    \includegraphics[trim=30mm 15mm 40mm 28mm,clip,width=1.0\columnwidth]{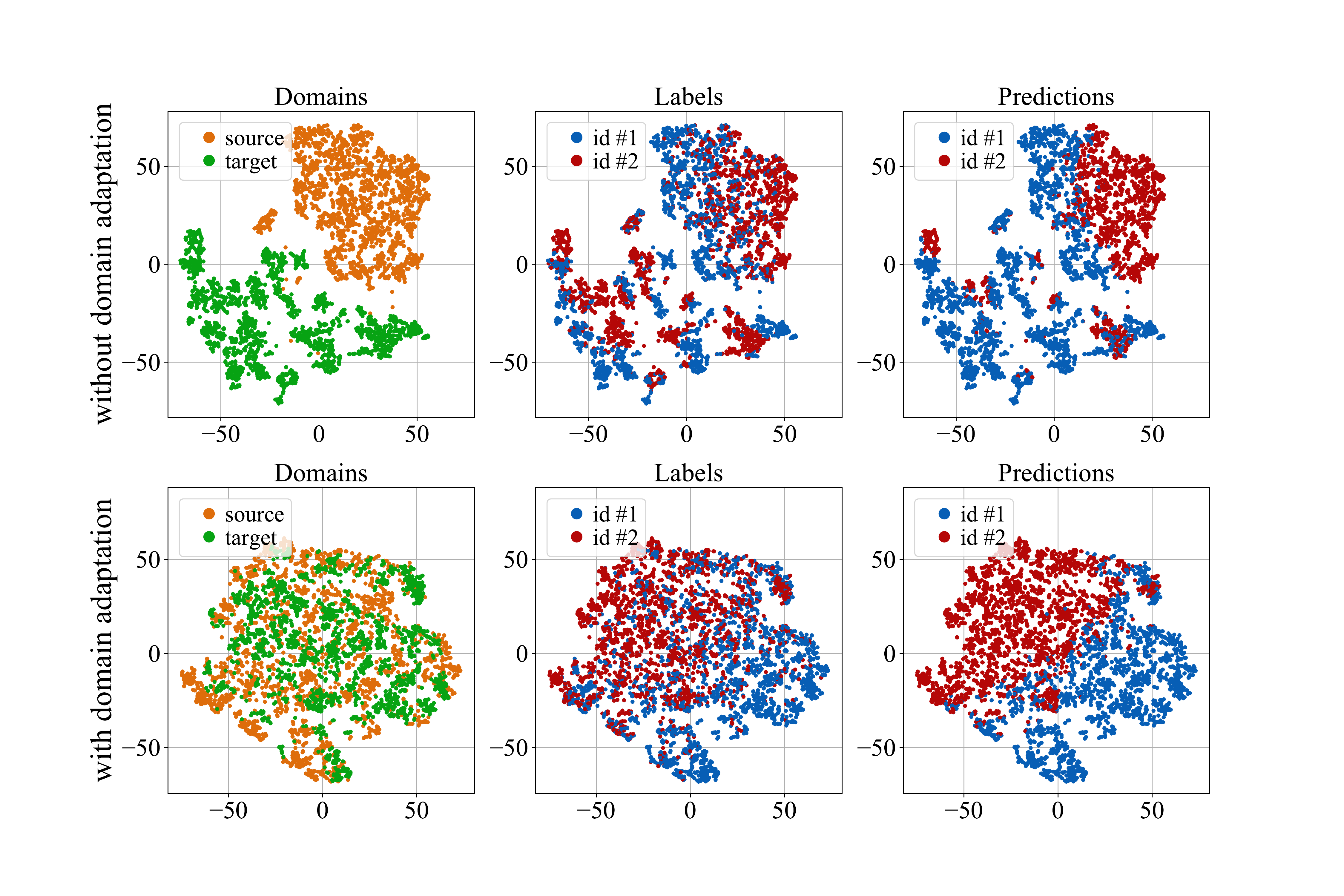}
    \caption{\small Illustration of results: feature distributions of the test data (extracted from the final layer before classification) in 2D using t-SNE dimensionality reduction technique~\cite{van2008visualizing}. \emph{First row:} Feature distributions of source and target domains before domain adaptation, where we observe that the network performs well on the source domain but not on the target domain. \emph{Second row:} Feature distributions after domain adaptation, which reflect that the network generalizes well to the target data as well.}
    \label{fig:feature-plots}
\end{figure}

We show our results in Figure~\ref{fig:feature-plots} where we visualize the feature distributions of the test samples before and after domain adaptation using t-SNE~\cite{van2008visualizing}. Before the domain adaptation (shown in the first row), we observe that the network trained on the source data produces two feature clusters for the source and target domains. Furthermore, ground truth labels of the source samples seem to be well-separated, which allows the network to achieve a classification accuracy of $75.80\%$ on the source domain, as illustrated in the predictions plot. Since the network does not see any target samples before the domain adaptation, it performs rather poorly on the target domain, achieving $62.70\%$ accuracy. After the domain adaptation (shown in the second row), we observe that the feature distributions of the source and target data are well-aligned, and ground truth labels for both domains seem to be well-separated, which allows the network to achieve a classification accuracy of $76.35\%$ on the target domain, as illustrated in the predictions plot. We also report average classification accuracies in Table~\ref{tab:average-accuracies} computed over $20$ independent trials using the same network and hyperparameters.

\setlength{\tabcolsep}{6pt}
\begin{table}[t]
 \vspace{1mm}
\begin{tabularx}{1.0\columnwidth}{*{3}{>{\centering\arraybackslash}X}}
\cmidrule(r){1-3}
\textbf{source} & \textbf{target (before adaptation)} & \textbf{target (after adaptation)} \\ \cmidrule(r){1-3}
$74.57\pm0.84$ & $59.67\pm9.26$ & $77.08\pm2.42$ \\  \bottomrule
\end{tabularx}
\vspace{1mm}
\caption{\small Average classification accuracies (in percentage) at different stages of our method, computed over $20$ independent trials.}
\label{tab:average-accuracies}
\end{table}

\begin{figure}[t]
\centering
    \begin{subfigure}[H]{0.85\columnwidth}
            \includegraphics[width=\linewidth]{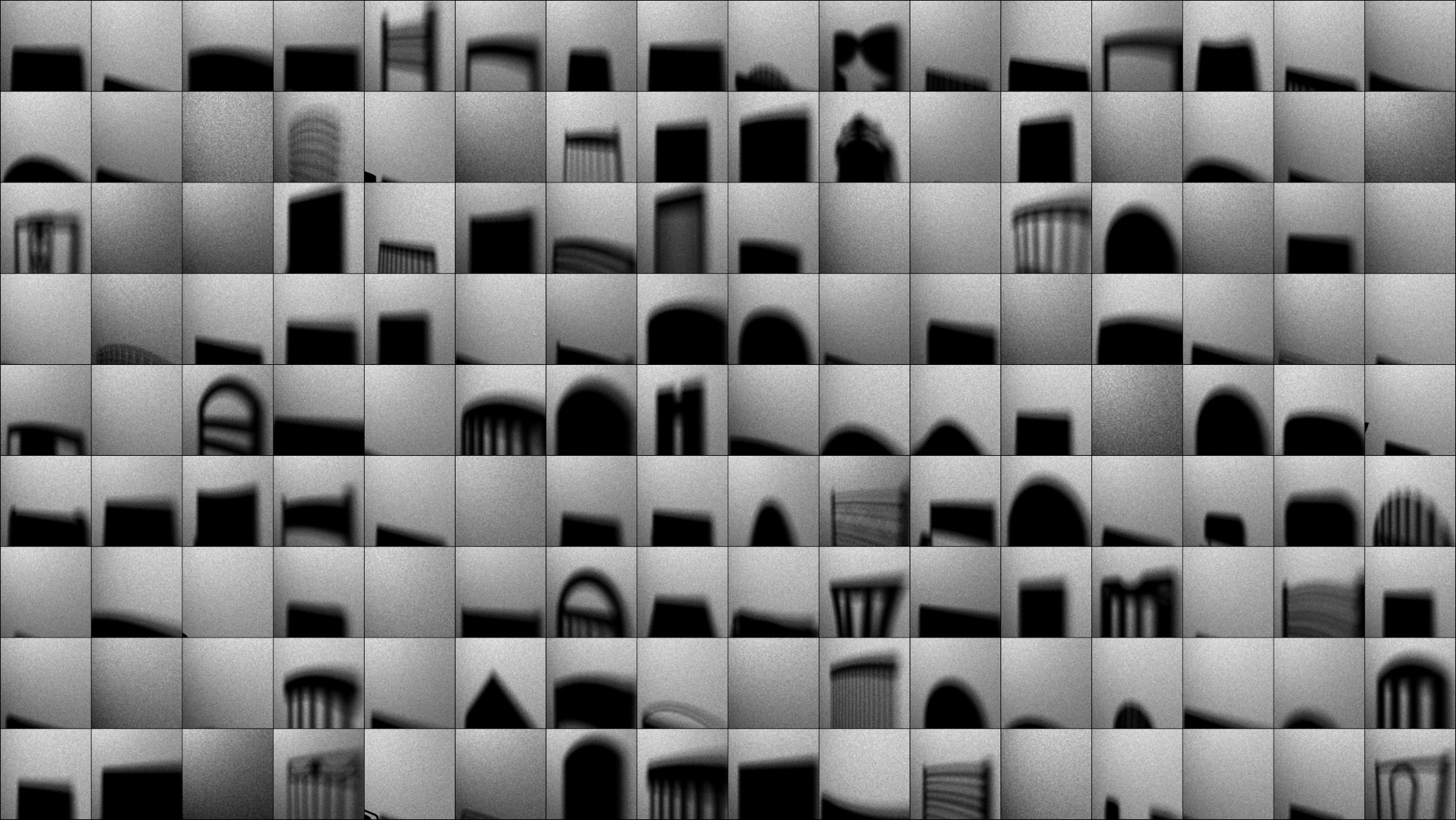}
            \caption{Incorrectly classified images}
            \vspace{0.2cm}
            \label{fig:incorrect-images}
    \end{subfigure}
    \hspace{0.1cm}
    \begin{subfigure}[H]{0.85\columnwidth}
            \includegraphics[width=\linewidth]{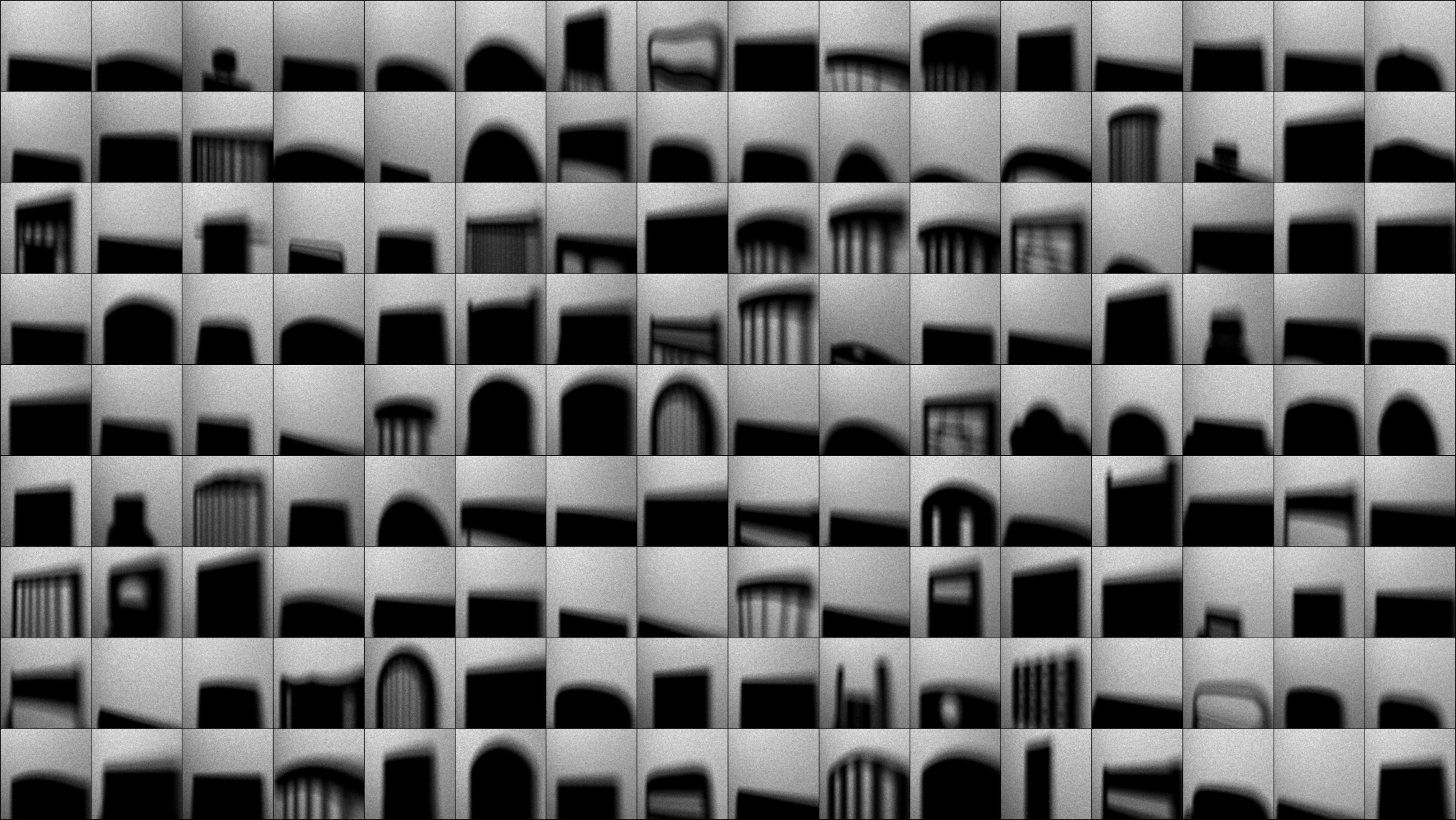}
            \caption{Correctly classified images}
            \label{fig:correct-images}
    \end{subfigure}
    \caption{\small Random samples of incorrectly and correctly classified images. Incorrectly classified images usually lack shadows (hence penumbrae) where most information appears to lie. In contrast, correctly classified images usually have large shadow areas.}
    \vspace{-1mm}
    \label{fig:incorrect-correct-images}
    \vspace{-3mm}
\end{figure}

\section{Discussion and Analysis}
\label{sec:discussion}

Having demonstrated that biometric information leakage \emph{can} occur, we now turn to understanding aspects of \emph{how} it occurs, by interpreting and analyzing  the behavior of our neural network classifier in various scene configurations. To achieve this, we analyze the results on the synthetic images for which we have access to the conditions under which they were rendered, such as occluder shape, head pose, and light source location. We analyze the samples on which the network fails or performs well, and the regions of the input that the network relies on the most by using interpretable machine learning tools referred to as saliency methods~\cite{adebayo2018sanity}.

We first investigate the influence of occluder shape and face appearance on the performance, where we compare all $484$ fail cases (which gives us $75.80\%$ accuracy on the source domain) with $484$ of the correctly classified images with the highest softmax probabilities. For the occluder shape analysis, we illustrate random samples from the incorrectly and correctly classified images in Figure~\ref{fig:incorrect-correct-images}. Here, it is observed that the incorrectly classified images usually lack shadows. Specifically,
defining black pixels (with zero intensity) in each image as \emph{umbra}, the umbrae cover $12.11\%$ of the incorrectly classified images on average, whereas they cover $21.95\%$ of the correctly classified images.

\begin{figure}[t]
\centering
    \includegraphics[width=1.0\columnwidth]{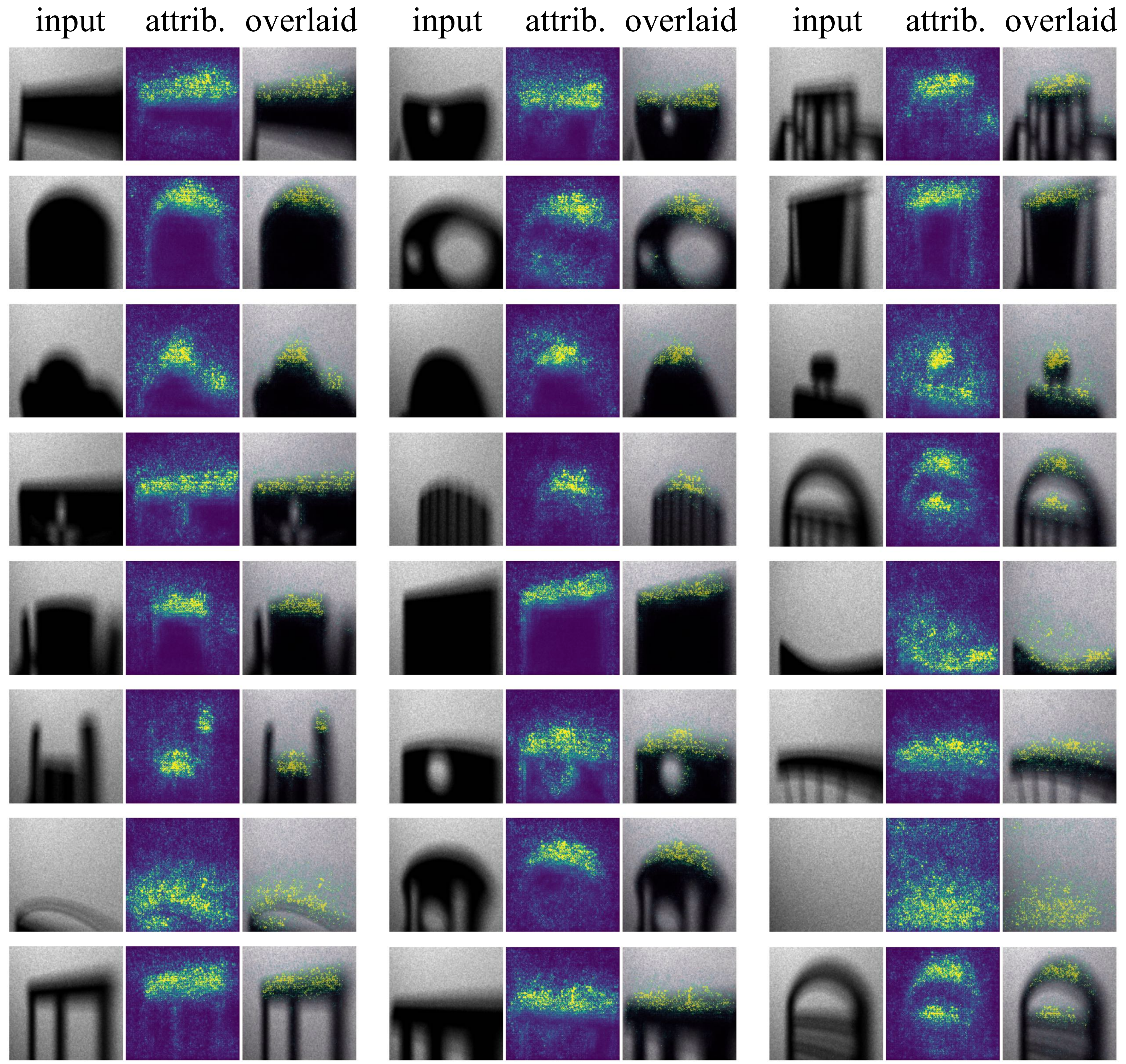}
    \caption{\small Image attributions extracted by the Integrated Gradients method \cite{sundararajan2017axiomatic}. We observe that the network is mostly sensitive to the penumbra regions, where most biometric information seems to lie.}
    \label{fig:integrated-gradients}
    \vspace{-3mm}
\end{figure}

The fact that the shadows appear to be crucial for inferring identities is consistent with the analysis of the resolving power of \emph{single edge occluders}~\cite{bouman2017turning, seidel2019corner, seidel2020two}. In our case, we use the resolving power of the \emph{edges of the occluder}, where the penumbra formed on the wall can be used to calculate 1D projections of the input face along the direction of the edges. In other words, our results suggest that the penumbrae contain the most useful information about the unknown identity present in the scene, and they are in fact where our network appears to rely on the most. In particular, we investigate which regions of the input images have more influence on the class predictions by employing a saliency method referred to as integrated gradients \cite{sundararajan2017axiomatic}. We illustrate several examples in Figure~\ref{fig:integrated-gradients}, where we show the original inputs and the image attributions for each input. We observe that the network is more sensitive to the penumbra regions compared to other parts of the image, which is in line with our previous observation that the penumbrae leak most of the sensitive information.

\section{Concluding Remarks}
We show that it is possible for biometric information of individuals to be inferred from indirect shadows cast by objects on diffuse surfaces. We analyze this largely overlooked optical phenomenon  first via a maximum likelihood analysis, which  shows that otherwise innocuous shadows can be exploited for reliable identity inference under representative scenarios. We further construct a method---representative of one that might be used in practice by an adversary---that demonstrates these vulnerabilities in real settings.  In particular, we use a learning-based approach that discovers hidden biometric cues in the indirect shadows by combining synthetic data training with unsupervised domain adaptation. Our synthetic data acquisition relies on a state-of-the-art 3D face reconstruction network, with which we obtain accurate 3D face models from only a single photograph of each identity. We show that our method achieves high accuracies in an identity classification task in real settings, and is robust to several variations in the scene, such as the shape of the occluding objects, lighting, head pose, and facial expressions. Our results suggest that the primary source of the biometric information leakage is the penumbra portions of the shadows, which we explain with the resolving power of occluding edges. Although the degree to which larger numbers of identities can be distinguished---and different types of biometric information can be extracted---remains to be investigated, our results make clear that biometric leakage occurs and that the information can be extracted by using existing tools and learning methodologies. Given that  indirect shadow phenomena is omnipresent, our results make a case for further investigation of the risks and an exploration of approaches to their mitigation.

At the same time, the extensions of our methodology could, in principle, facilitate applications that would have positive societal impacts. For instance, such extensions would be useful in certain security and surveillance applications, or in identity recognition tasks that require no storage or observation of any sensitive information about the identities, enabling face recognition without taking any photos of the individuals. These, too, warrant further investigation.

{\small
\bibliographystyle{ieee_fullname}
\bibliography{bibl}
}

\appendix
\twocolumn[\LARGE \centering \textbf{Supplementary Material} \vspace{5mm}]

\section{Implementation Details}
In the following, we explain our data generation processes, the implementation of our maximum likelihood-based learning algorithm, and the training procedure of our neural network classifier in detail.

\subsection{Maximum Likelihood Analysis}

\textbf{Dataset generation.} According to the coordinate system definition shown in Figure 5(a), we have the following configurations and variations in the face dataset for each of the $M=16$ identities. Hereafter, with a slight abuse of notation, we denote a point in the 3D-space by $(x, y, z)$, where the unit of measure is meters.
\begin{itemize}
    \item We fix the position of the faces at $(1.65, 0.0, 1.15)$.
    
    \item We vary the facial expressions by sampling the expression coefficients from $\mathcal{N}(\mathbf{0}_{k_\mathrm{exp}}, c\, \mathbf{I}_{k_\mathrm{exp}})$, which changes the face shape according to Equation (1). Here, $c$ controls the amount of variations, which we set as $c = 0.5$.
    
    \item We rotate the faces around $y$-- and $z$--axes, which we refer to as elevation and azimuth, respectively. We sample both elevation and azimuth uniformly from $[-15^{\circ}, 15^{\circ}]$.
    
    \item We simulate a white spotlight directed to the face. We sample its location uniformly along the line connecting $(0.15, -0.5, 1.50)$ and $(0.15, 0.5, 1.50)$.
    
\end{itemize}

We render face images of resolution $128\times128$ with Mitsuba2~\cite{nimier2019mitsuba} using $100$ samples per pixel. Rendering one image takes $\sim\!100$  miliseconds on NVIDIA GeForce RTX 2080 Ti GPU. We simulate a rectangular occluder with a $512 \times 512$ image that contains a square with a diagonal length of $400$ pixels. After proper scaling in pixel values to ensure the conservation of energy~\cite{ajjanagadde2019near, yedidia2018analysis}, we convolve this image with the rendered face images to obtain the shadow images, which we downsample to $128\times128$ resolution. Hence, we have $ n = 128^2 = 16384$, i.e, each observation in the dataset is a $16384$-dimensional vector.

\textbf{Experiments.} We analyze the ML algorithm for different SNR levels ranging from $-35$ dB to $80$ dB with step size $5$ dB. For each data point, running the algorithm for $M=16$ identities takes $\sim\!150$ minutes. For each data point shown in Figure 6, we compute the accuracies by averaging the results over $5$ independent trials.

A particular setting that requires careful attention is when the noise variance ${\sigma}^2$ is sufficiently small, which makes the covariance matrices $\{\mathbf{Q}^m\}$ nearly singular. This makes the ``tail" eigenvalues of the sample covariance matrix $\mathbf{\hat{Q}}^{m}$ close to $0$, causing numerical instabilities during inversion. To resolve this, if the estimated noise variance $\widehat{\sigma}^2$ drops below some chosen threshold $\sigma^2_{\mathrm{th}}$, we work with the \emph{pseudoinverse} of the covariance matrix as \smash{$(\mathbf{\hat{Q}}^{m})^{-1} \triangleq \mathbf{U}^m (\mathbf{{\Lambda}}^m)^+ (\mathbf{U}^m)^T  $} where $(\mathbf{{\Lambda}}^m)^+ \triangleq \mathrm{diag}(1/\lambda_1^m, \dots, 1/\lambda_p^m, 0, \dots, 0)$. Here, $\lambda_p^m$ is the smallest eigenvalue larger than a refined threshold $\bar{\sigma}^2_{\mathrm{th}} \triangleq \max(\sigma^2_{\mathrm{th}}, k \widehat{\sigma}^2)$, which is heuristically chosen to ensure the continuity of the algorithm between numerically singular and nonsingular covariance matrix regimes. In our experiments, we set $\sigma^2_{\mathrm{th}} = 10^{-6}$ and $k = 5$.

\subsection{Neural Network Classifier}

\textbf{Dataset Generation.} According to the same coordinate system definition, we have the following configurations and variations in the synthetic data, which we use to train our neural network classifier.
\begin{itemize}
    \item As before, we vary the facial expressions by sampling the expression coefficients from $\mathcal{N}(\mathbf{0}_{k_\mathrm{exp}}, c\, \mathbf{I}_{k_\mathrm{exp}})$ with $c=0.5$.

    \item We rotate the faces around $y$-- and $z$--axes, and sample both elevation and azimuth uniformly from $[-30^{\circ}, 30^{\circ}]$. Here, positive angles indicate clockwise rotations with respect to the $xz$-- and $xy$--planes.

    \item We sample the position of the face uniformly along the line connecting $(1.55, 0.0, 1.15)$ and $(1.75, 0.0, 1.15)$. That is, face position varies along the $x$--axis as variations in other axes are accounted for in the data augmentation step, where the final images are randomly cropped.
        
    \item We simulate a white spotlight directed to the face. We sample its location uniformly along the line connecting $(0.15, -1.0, 1.50)$ and $(0.15, 1.0, 1.50)$.
    
    \item Occluders are located $0.7$ meters from the wall and situated on the ground, where we measure the distance from the center of mass of the occluder. 
    
\end{itemize}

\textbf{Training details.} We train our classification network with the synthetic data for $30$ epochs, using cross entropy loss and Adam optimizer \cite{kingma2014adam} with a learning rate of $0.0001$. We augment the training data by flipping the images randomly, resizing them to $280\times280$ resolution and randomly cropping a $224\times224$ patch from these images. At test time, we resize the images to $280\times280$ resolution and center-crop the $224\times224$ patch from them. In our experiments, we pick the epoch with the highest test accuracy, and use the network at that epoch as our baseline, on which we apply domain adaptation by updating the batch normalization statistics.

\section{Fundamental Limitations} 

The performance of identity classification systems from shadows (like the one that we present in this work) is fundamentally influenced by several factors such as the similarities between identities of interest; the amount of variations covered in head poses, facial expressions, lighting conditions, occluding objects; and other less trivial factors.

\begin{figure}[t]
\centering
     \includegraphics[width=1.0\columnwidth]{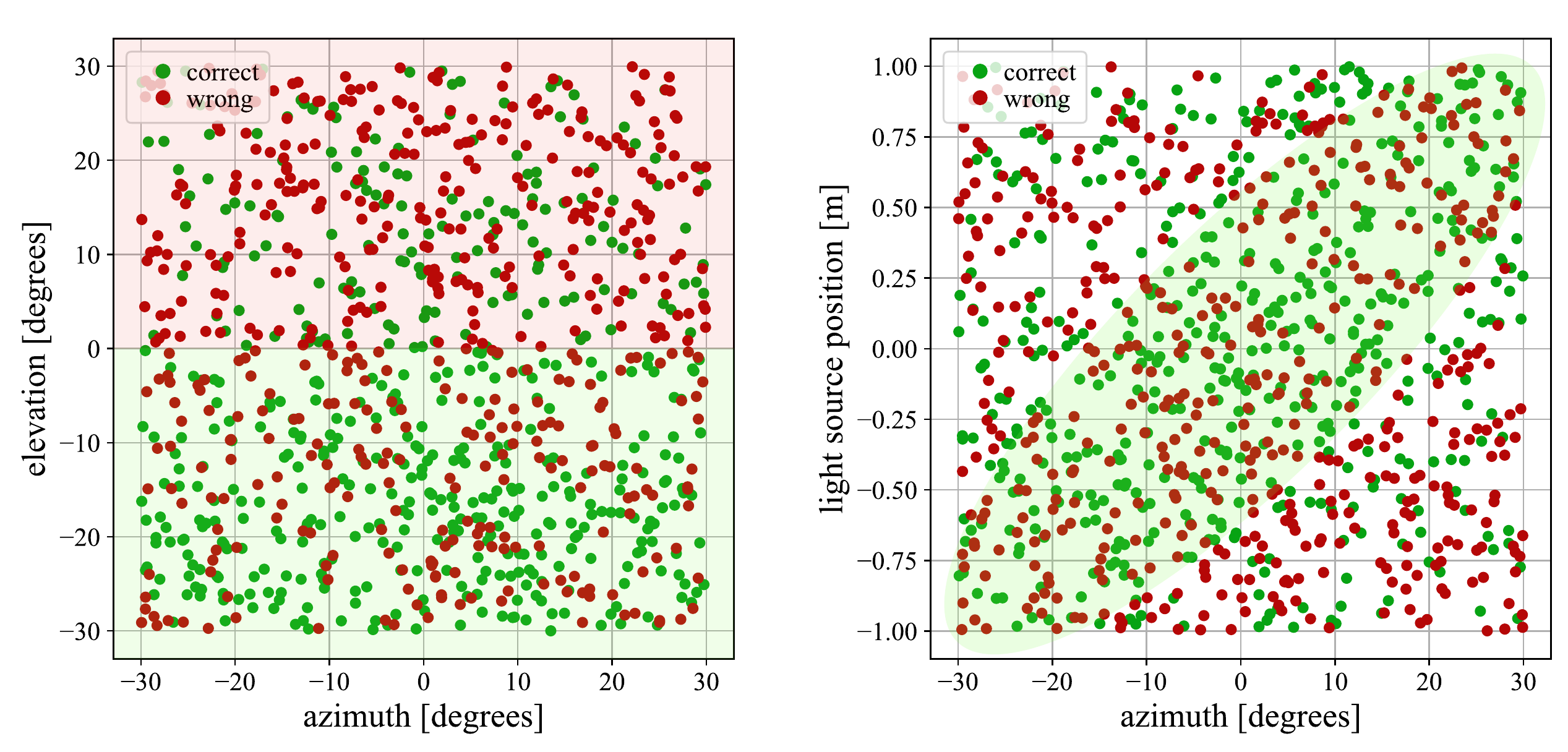}
    \caption{Correctly and incorrectly classified examples depending on azimuth, elevation and light source position. We observe that faces with lower elevations and less cast shadows are more likely to be classified correctly.}
    \label{fig:incorrect-correct-plots}
\end{figure}

As an illustrative example, it is expected that the faces under extreme head poses and lighting conditions are more likely to be classified incorrectly. To support this, we investigate the effect of the face appearance on the results by analyzing the impact of the head pose and light source location on the predictions. We illustrate our findings in Figure~\ref{fig:incorrect-correct-plots}, where we show elevation-azimuth and light source position-azimuth plots for correctly and incorrectly classified examples. In the left plot, we observe that the elevation has an evident impact on classification performance, where faces with higher elevation are more likely to be misclassified. This can be explained by our scene geometry shown in Figure~ 5(a), where a more direct view of the face is reflected on the wall when the elevation is low. Taking the averages over all samples shown in the plot, incorrectly classified examples have an average elevation of $+2.83$ degrees whereas correctly classified examples have an average of $-6.73$ degrees. In the right plot, we observe a positive correlation (a Pearson correlation of $0.22$) between the light source position (measured along the $y$-axis) and the azimuth for correctly classified examples, for which the faces are illuminated with lower incidence angles. That is, faces with less cast shadows are more likely to predicted correctly.

\end{document}